\documentclass[10pt,twocolumn,letterpaper]{article}

\usepackage{wacv}
\usepackage{times}
\usepackage{epsfig}
\usepackage{graphicx}
\usepackage{amsmath}
\usepackage{amssymb}
\usepackage{algorithm}
\usepackage{algorithmic}

\usepackage{caption}
\usepackage{subcaption}

\def\wacvPaperID{529} 

\wacvfinalcopy 

\ifwacvfinal
\def\assignedStartPage{9876} 
\fi

\ifwacvfinal
\usepackage[breaklinks=true,bookmarks=false]{hyperref}
\else
\usepackage[pagebackref=true,breaklinks=true,colorlinks,bookmarks=false]{hyperref}
\fi

\ifwacvfinal
\else
\fi

\begin{document}

\title{AVGZSLNet: Audio-Visual Generalized Zero-Shot Learning by Reconstructing Label Features from Multi-Modal Embeddings}


\author{
Pratik Mazumder$^{\dagger}$ \hspace{1cm}Pravendra Singh$^{\dagger}$ \hspace{1cm}Kranti Kumar Parida$^{\dagger}$ \hspace{1cm}Vinay P. Namboodiri$^{\dagger \ast}$\\
$^{\dagger}$Department of Computer Science and Engineering, IIT Kanpur, India\\
$^{\ast}$University of Bath, United Kingdom \\
{\tt\small \{pratikm, psingh, kranti\}@iitk.ac.in, vpn22@bath.ac.uk}
}

\maketitle

\begin{abstract}
In this paper, we propose a novel approach for generalized zero-shot learning in a multi-modal setting, where we have novel classes of audio/video during testing that are not seen during training. We use the semantic relatedness of text embeddings as a means for zero-shot learning by aligning audio and video embeddings with the corresponding class label text feature space. Our approach uses a cross-modal decoder and a composite triplet loss. The cross-modal decoder enforces a constraint that the class label text features can be reconstructed from the audio and video embeddings of data points. This helps the audio and video embeddings to move closer to the class label text embedding. The composite triplet loss makes use of the audio, video, and text embeddings. It helps bring the embeddings from the same class closer and push away the embeddings from different classes in a multi-modal setting. This helps the network to perform better on the multi-modal zero-shot learning task. Importantly, our multi-modal zero-shot learning approach works even if a modality is missing at test time.  We test our approach on the generalized zero-shot classification and retrieval tasks and show that our approach outperforms other models in the presence of a single modality as well as in the presence of multiple modalities. We validate our approach by comparing it with previous approaches and using various ablations.

\end{abstract}

\vspace{-10pt}
\section{Introduction}
Deep learning methods have become extremely popular in language processing as well as in computer vision tasks on images, videos, and sounds. Traditional deep learning methods involve training the network on massive amounts of data, with each category/class having a large number of data points. However, in many real-life settings, data may not be available for all the classes. In such cases, the standard deep learning training methods cause the model to overfit to the classes seen during training and fail miserably in classifying the test data points from classes not seen during training. Humans, on the other hand, can perform much better in such a setting. Given the description of a category/class, such as a cat, a person can identify pictures of cats with a high probability of success. Further, if the person has already seen similar animals such as tigers, leopards, he/she will be even better at identifying cats. We want deep learning models to also work well in similar settings. Zero-shot learning is an area of machine learning that deals with this problem setting. 

Zero-shot learning involves specialized training of networks in order to enable them to classify unseen classes at test time using only basic class information such as class name, description, or attributes \cite{frome2013devise, norouzi2013zero, xian2018feature}. Zero-shot learning methods help networks to learn good semantic representations such that they can easily transfer the knowledge gained from seen classes to the unseen classes. In the standard zero-shot learning setting, at test time, the data is only from the unseen classes \cite{lampert2013attribute}. Recently, a more practical setting, namely generalized zero-shot learning, is being used, where the test data comes from both seen and unseen classes \cite{kumar2018generalized}. This setting is harder because the model is biased towards the seen classes.

Researchers have extensively applied zero-shot learning to images, videos, and sounds. Very recently, the authors in \cite{parida2020coordinated} dealt with a multi-modal setup where each data point consisted of a video and corresponding audio. It applies zero-shot learning to this setup and shows how audio information can help to better classify videos under the zero-shot setting, e.g., honking of a car even if it is not visible (occluded) in the video. 

We propose a novel approach to address the generalized multi-modal zero-shot learning problem in the audio-visual setting. We use a cross-modal decoder to optimize the audio and video embeddings in such a way that we can reconstruct the class label text feature from them. This forces the projection networks to include class-level information in the audio and visual embeddings. We also use a composite triplet loss with audio, video, and text anchors to force audio/video embeddings to move closer to the label text embedding of their corresponding classes and farther away from other classes. We provide a detailed description of our method in Sec. \ref{sec:method}.

At test time, the data points (audio or video or both) get projected to our learned embedding space where we perform the nearest neighbor classification. We predict the class label embedding that is closest to the multi-modal embedding of the test data point as the output class. We perform experiments for the generalized zero-shot classification and retrieval tasks on the AudioZSL~\cite{parida2020coordinated} dataset to show the efficacy of our method. We validate the components of our method through extensive ablation experiments. We compare our approach with state-of-the-art methods.

Our contributions can be summarized as follows:

\begin{itemize}
    \item We propose a novel method to address the Audio-Visual Generalized Zero-Shot Learning problem by using a cross-modal decoder and a composite triplet loss.
    \item We experimentally show that our method outperforms the state-of-the-art method for both generalized zero-shot audio-visual classification and retrieval.
    \item We empirically show that our method performs well even when only audio or video data is present for the test data point.
\end{itemize}
\section{Related Works}
\textbf{Zero-Shot Learning:}
Zero-shot learning \cite{socher2013zero, xian2016latent, akata2015evaluation, xian2018feature, chao2016empirical, frome2013devise, romera2015embarrassingly, lampert2013attribute, norouzi2013zero, akata2015label, kumar2018generalized, kodirov2017semantic} involves training a network in a specialized way so that it can reasonably classify unseen classes. A common training approach is to push the feature embeddings of data close to the semantic embedding of their class. Semantic embedding of their class are generally obtained by using the class label text \cite{frome2013devise, norouzi2013zero, xian2018feature} or description/attributes \cite{lampert2013attribute, romera2015embarrassingly, xian2018feature}.
Some methods \cite{akata2015label, xian2016latent, akata2015evaluation} use both kind of class information. 
At test time, the query data can be from the unseen classes only \cite{lampert2013attribute} or from both seen and unseen classes. The latter setting is called the generalized zero-shot learning \cite{chao2016empirical}.

\textbf{Audio-Visual Learning:}
Several works combine audio and video data to improve network performance in various tasks such as audio-visual correspondence learning \cite{aytar2016soundnet, owens2016ambient, arandjelovic2017look, owens2018audio}, audio-visual source separation \cite{zhao2018sound, gao2018learning} and others. The method proposed in \cite{owens2016ambient} utilizes self-supervision to detect the temporal alignment between audio and video. Features learned from this setting can be applied to several downstream tasks like source localization and action recognition. The authors in \cite{ephrat2018looking} propose to use an audio-visual setting to train a model to perform speaker-independent speech source separation. The work in \cite{zhao2018sound} uses self-supervision to perform pixel-level audio source localization. The authors in \cite{parida2020coordinated} propose CJME and experimentally show how adding audio information to video data can improve the model performance for zero-shot classification and retrieval.

\section{Proposed Method}\label{sec:method}
\begin{figure}[t]
  \centering
  \includegraphics[width=\linewidth]{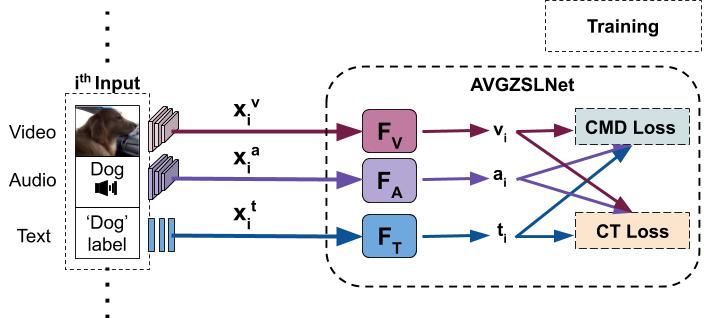}
  \caption{AVGZSLNet Training. Input data is in the form of tuples containing video, audio and class label. Pretrained networks for video/audio/text extract the video features $x_i^v$, audio features $x_i^a$ and text features $x_i^t$ from the video, audio and class label data, respectively. Projection networks $F_V,F_A$ and $F_T$ project video, audio and text features to the embeddings $v_i,a_i$ and $t_i$ respectively. The complete network (AVGZSLNet) is optimized on a combination of the cross-modal decoder (CMD) loss and the composite triplet (CT) loss.}
  \label{fig:train}
  \vspace{-15pt}
\end{figure}

\subsection{Problem Setting}
In the audio-visual generalized zero-shot learning problem, each training data point consists of a tuple containing audio, video, and class label from the seen classes $S$. At the time of testing, the network has to predict the class label for the query data. The query data can belong to both seen ($S$) and unseen ($U$) classes. Query data can have either audio or video or both. The objective is to predict the output class for each query data by using the text embedding of each class label and the audio/video/both embedding of the query data.

\subsection{Method Description}
We describe our proposed Audio-Visual Generalized Zero-Shot Learning Network (AVGZSLNet) in detail in this section.

Each input data point $i$ consists of video, audio, and class label ($y_i$). We use pre-trained networks for video, audio, and text to extract the corresponding video features $x_i^v$, audio features $x_i^a$, and class label text features $x_i^t$ for the $i^{th}$ data point. During training, for each mini-batch, we randomly sample multiple tuples of multi-modal features. Each tuple consists of audio, video, and class label text features. Let $(x^a_p, x^v_p, x^t_p)$ and $(x^a_q, x^v_q, x^t_q)$ be two tuples of multi-modal features, where $x^a_p,x^a_q$ represent the audio features; $x^v_p,x^v_q$ represent the video features; $x^t_p,x^t_q$ represent the class label text features for the class labels $y_p$ and $y_q$ ($y_p\neq y_q$). Our method is trained on such tuples. For training, we first obtain the embeddings for audio, video, and label text features using the embedding/projection networks (Fig. \ref{fig:train}).
\begin{equation}\label{eq:ext1}
    a_p,v_p,t_p = F_A(x^a_p),F_V(x^v_p),F_T(x^t_p)
\end{equation}
\begin{equation}\label{eq:ext2}
    a_q,v_q,t_q = F_A(x^a_q),F_V(x^v_q),F_T(x^t_q)
\end{equation}
Where $F_A$, $F_V$, $F_T$ are embedding/projection networks for audio, video and text features. $a_p, v_p, t_p$ are the generated audio, video and text embeddings for $x^a_p, x^v_p, x^t_p$. $a_q, v_q, t_q$ are the generated audio, video and text embeddings for $x^a_q, x^v_q, x^t_q$.

Our goal is to train these embedding networks in such a way that the audio and video embeddings are close to the class label text embedding of the same class and are away from the class label text embeddings of different classes. To achieve this objective, we train the embedding networks on a combination of cross-modal decoder loss and composite triplet loss (Fig. \ref{fig:train}). We explain both the losses in detail below.

We would like to point out that, in our method, the embedding and feature terms are not interchangeable. Audio ($x_i^a$) / video ($x_i^v$) / class label text ($x_i^t$) features are extracted by the pre-trained network from the audio/video/class label text data. Audio ($a_i$) / video ($v_i$) / class label text ($t_i$) embeddings are generated by the embedding/projection networks from the audio ($x_i^a$) / video ($x_i^v$) / class label text ($x_i^t$) features.

\begin{figure}[t]
  \centering
  \includegraphics[width=\linewidth]{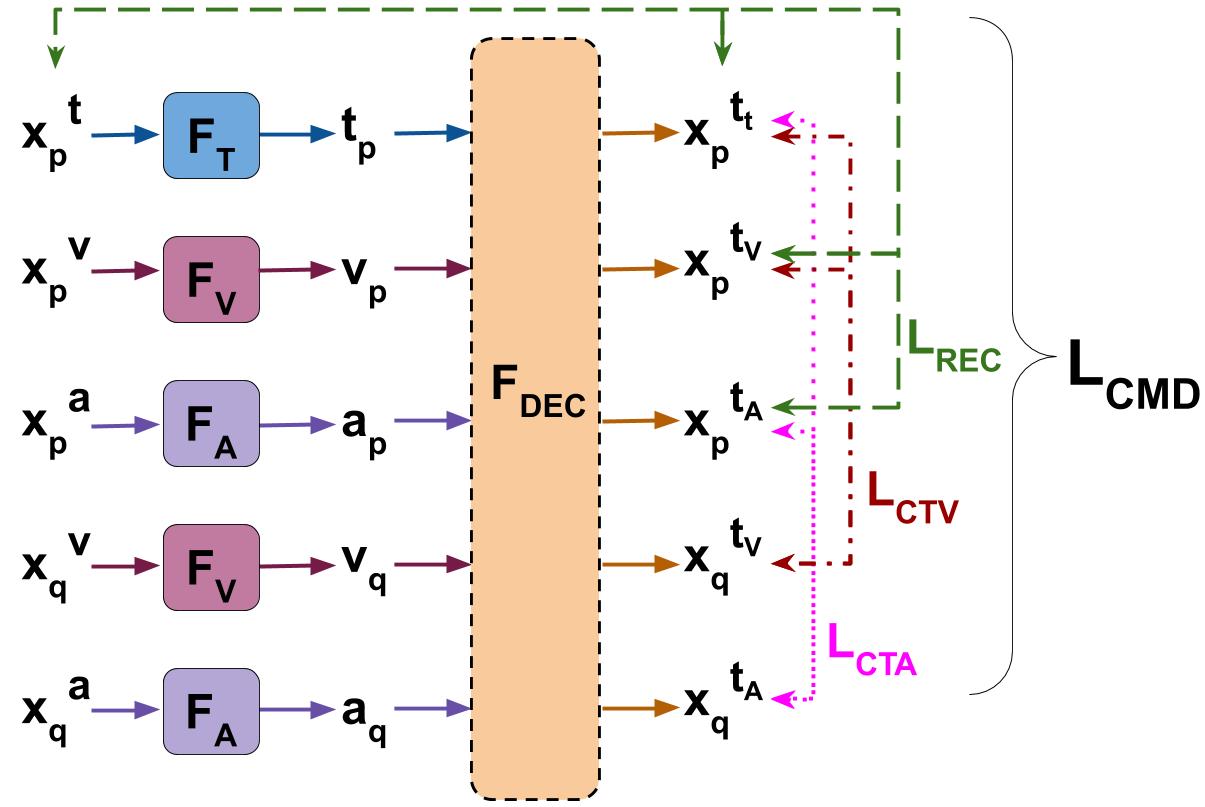}
  \caption{\textbf{Proposed Cross-Modal Decoder}. $(x^a_p, x^v_p, x^t_p)$ and $(x^a_q, x^v_q, x^t_q)$ are two data points with class labels $y_p$ and $y_q$ ($y_p\neq y_q$) from the training set where $x^a_p,x^a_q$ represent audio features; $x^v_p,x^v_q$ represent video features; and $x^t_p,x^t_q$ represent text label features. $F_A$, $F_V$, $F_T$ are embedding networks for audio, video and text features; $a_p, v_p, t_p$ represent the audio, video and text embeddings generated by the projection/embedding networks from $x^a_p, x^v_p, x^t_p$ respectively; and $a_q, v_q, t_q$ represent the audio, video and text embeddings generated by the projection/embedding networks from $x^a_q, x^v_q, x^t_q$ respectively. $F_{DEC}$ projects each type of embeddings to the text feature of the data point. $x_p^{t_t},x_p^{t_v},x_p^{t_a}$ refer to the reconstructed text feature from $a_p, v_p, t_p$. $x_q^{t_t},x_q^{t_v},x_q^{t_a}$ refer to the reconstructed text feature from $a_q, v_q, t_q$. The $L_{DEC},L_{CTV},L_{CTA}$ losses are applied to these outputs to obtain the cross-modal decoder loss $L_{CMD}$.}
  \label{fig:aeloss}
  \vspace{-10pt}
\end{figure}

\subsubsection{Cross-Modal Decoder Loss} \label{sec:cmd}
In order to have a better cross-modal semantic alignment, the audio or video embeddings should contain the class label information encoded in some form. Therefore, projection networks should produce audio and video embeddings that contain information about the text features of that class. To achieve this, we propose a cross-modal decoder (Fig.~\ref{fig:aeloss}) that is trained to reconstruct the class label text features from the video embeddings, audio embeddings and text embeddings separately. This will force the embedding networks to include the class information in the video, audio embeddings (Eq.~\ref{eq:dec}). We also use a triplet loss to ensure that the reconstructed text feature from the text embedding is close to the reconstructed text features from the audio and video embeddings of the same class (Eqs.~\ref{eq:dec_audio_trip}, \ref{eq:dec_video_trip}).
\begin{multline}\label{eq:dec}
       L_{REC}(a_p,v_p,t_p,x^t_p)= d(F_{DEC}(t_p),x^t_p)  \\+ d(F_{DEC}(a_p),x^t_p)   + d(F_{DEC}(v_p),x^t_p) 
\end{multline}
\begin{multline}\label{eq:dec_audio_trip}
       L_{CTA}(a_p,t_p,x^t_p,a_q)= [d(F_{DEC}(t_p),F_{DEC}(a_p))-\\d(F_{DEC}(t_p),F_{DEC}(a_q))+\delta]_{+}  
\end{multline}
\begin{multline}\label{eq:dec_video_trip}
       L_{CTV}(v_p,t_p,x^t_p,v_q)= [d(F_{DEC}(t_p),F_{DEC}(v_p))-\\d(F_{DEC}(t_p),F_{DEC}(v_q))+\delta]_{+}  
\end{multline}
\begin{multline}\label{eq:ae}
       L_{CMD} = L_{REC} + L_{CTA} + L_{CTV}
\end{multline}
Where $F_{DEC}$ is the shared decoder network for text, audio and video embeddings that is used to re-construct the label text features from the video, audio and text embeddings; $L_{REC}$ is the reconstruction loss for the text features generated from the audio, video and label text embeddings; $L_{CTA}$ is the triplet loss with text feature reconstructed from text embedding as the anchor, the text feature reconstructed from $a_p$ as the positive input and the text feature reconstructed from $a_q$ as the negative input; $L_{CTV}$ is the triplet loss with text feature reconstructed from text embedding as the anchor, the text feature reconstructed from $v_p$ as the positive input and the text feature reconstructed from $v_q$ as the negative input; $L_{CMD}$ is the cross-modal decoder loss, $\delta$ is the margin value, $d$ is the distance metric. We use mean square error (MSE) for the distance metric.

In the process of minimizing $L_{CMD}$, the projection networks learn to produce audio and video embeddings that contain the class information. We experimentally show that this improves the performance of the network.

\begin{figure}[t]
  \centering
  \includegraphics[width=\linewidth]{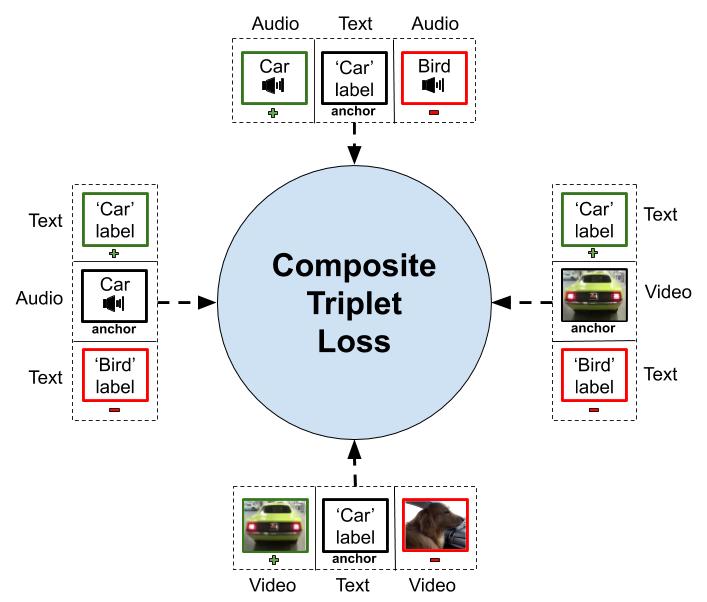}
  \caption{Composite triplet loss. Four types of triplets are formed using either the label text embedding, audio embedding and video embedding as the anchor.}
  \label{fig:triploss}
 \vspace{-10pt}
\end{figure}

\subsubsection{Composite Triplet Loss}\label{sec:ct}
In order to help the network bring audio and video embeddings closer to the text embedding of their corresponding class and push embeddings of different classes further away, we use a composite triplet loss. Triplet loss generally involves comparing a base input/anchor to a positive input and a negative input. We formulate multiple cross-modal triplet losses by using either text, audio and video data as the anchor input (Fig.~\ref{fig:triploss}, Eqs.~\ref{eq:lta}, \ref{eq:lat}, \ref{eq:ltv}, \ref{eq:lvt}).
\begin{equation}\label{eq:lta}
    L_{TA}(a_p,t_p,a_q,t_q) = [d(a_p,t_p)-d(a_q,t_p)+\delta]_{+} 
\end{equation}
\begin{equation}\label{eq:lat}
    L_{AT}(a_p,t_p,a_q,t_q) = [d(t_p,a_p)-d(t_q,a_p)+\delta]_{+}
\end{equation}
\begin{equation}\label{eq:ltv}
    L_{TV}(v_p,t_p,v_q,t_q) = [d(v_p,t_p)-d(v_q,t_p)+\delta]_{+}
\end{equation}
\begin{equation}\label{eq:lvt}
    L_{VT}(v_p,t_p,v_q,t_q) = [d(t_p,v_p)-d(t_q,v_p)+\delta]_{+}
\end{equation}
\begin{equation}\label{eq:ct}
    L_{CT} =  L_{TV} + L_{VT} + L_{TA} + L_{AT}
\end{equation}
Where $L_{TA}$ and $L_{TV}$ are the triplet loss on audio and video embeddings respectively with the class label text embedding as the anchor. $L_{AT}$ and $L_{VT}$ are the triplet loss on the class label text embeddings with the audio and video embeddings as the respective anchors. $L_{CT}$ denotes the total composite triplet loss.

\subsubsection{Total Loss}
We have defined the cross-modal decoder loss ($L_{CMD}$) in Sec. \ref{sec:cmd} and the composite triplet loss ($L_{CT}$) in Sec. \ref{sec:ct}. Therefore, the full loss function can be defined as follows:
\begin{equation}\label{eq:full}
    L = L_{CMD} + L_{CT}
\end{equation}
\subsubsection{Testing}
\begin{figure}[t]
  \centering
  \includegraphics[width=\linewidth]{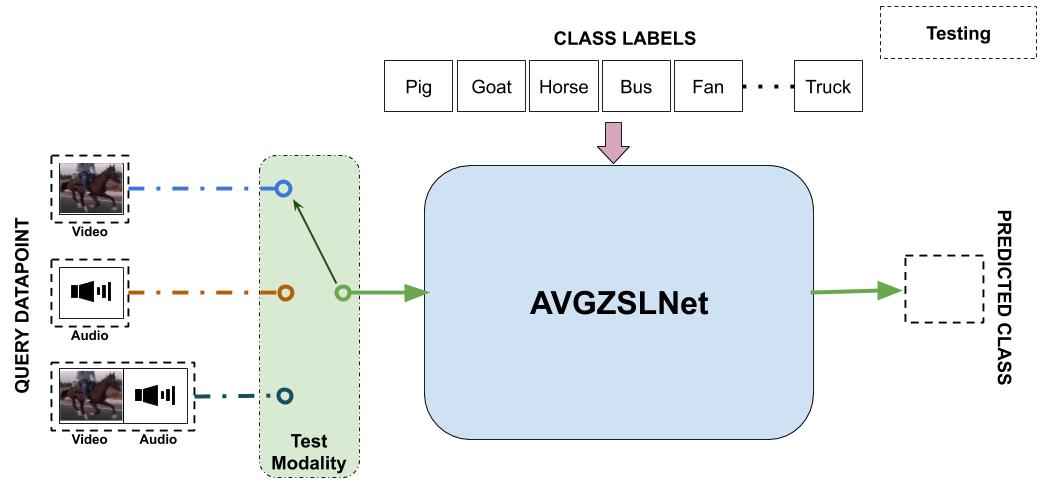}
  \caption{AVGZSLNet Testing. Query data can have either audio or video or both modalities. The trained projection networks are used to obtain the audio and/or video embeddings of the query data and the class label text embeddings of all the seen and unseen classes. The nearest class label text embedding to the audio or video or both embedding is predicted as the output class. The query data can be from both seen and unseen classes.}
  \label{fig:test}
  \vspace{-10pt}
\end{figure}
During testing (Fig. \ref{fig:test}), we first obtain the label text embedding for all the classes using $F_T$. Next, for each query data point, we obtain the embeddings for audio, video features using $F_A$, and $F_V$, respectively. Then, we calculate the distance of the audio and video embeddings of each query data point from every class label text embedding. We find the mean of the audio and video distances to obtain the final distance between the query and each class label text embeddings. Using this distance, we predict the nearest class text embedding as the output class. If one of the modalities (audio or video) is missing, then we calculate the distance only for the available modality and find the nearest class label text embedding. We provide further details on the testing modality in Sec. \ref{sec:testmod}.

\section{Experiments}
\subsection{Dataset}
We use the AudioSetZSL dataset \cite{parida2020coordinated} for the audio-visual generalized zero-shot learning. It is a subset of the AudioSet \cite{gemmeke2017audio} dataset. AudioSet consists of segments from YouTube videos along with the audio. The videos have different audio event labels, such as different types of human/animal/environmental sounds and sounds of musical instruments. Each video segment can have multiple audio event labels, and there are 527 labels in total.

In order to select videos for AudioSetZSL\footnote{https://github.com/krantiparida/AudioSetZSL} from AudioSet, the authors in \cite{parida2020coordinated} discard highly correlated classes to obtain a total of 33 classes.  Next, they discard all video segments containing more than one of the selected classes to create a multi-class classification setting. Finally, they select 1,56,416 video segments from the AudioSet dataset for the AudioSetZSL dataset. All the videos are of the same duration as defined in AudioSet \cite{gemmeke2017audio}. Statistics on the AudioSetZSL dataset are given in \cite{parida2020coordinated}. The 33 classes consist of 23 seen and 10 unseen classes. The unseen classes have minimal overlap with the Kinetics dataset \cite{carreira2017quo} training classes. 

For the audio-visual zero-shot learning task, we train the network on the seen class training examples and evaluate it on the test examples from both seen and unseen classes.

\subsection{Implementation Details}
We use the audio network proposed in \cite{kumar2018knowledge} as the pre-trained network to extract audio features from the audio data. We train it on the audio data spectrogram in the trainset of AudioSetZSL. We extract the audio features from this network after the seventh convolution layer and obtain a 1024 dimension vector by averaging. For video features, we use an inflated 3D CNN network pre-trained on the Kinetics action recognition dataset \cite{carreira2017quo}. The video feature is obtained in the same way, i.e., from the layer before the classification layer and averaged to a 1024 dimension vector. For the text features, we use the word2vec network that has been pre-trained on the Wikipedia dataset to obtain 300 dimension features \cite{mikolov2017advances}. We use a $\delta$ value of 1.

The projection network for audio and video embeddings are 2 layer fully connected networks. The text embedding network is a 1 layer fully connected network. The output dimension of all the 3 projection networks is 64. The cross-modal decoder network in our method is a 2 layer fully connected network with an output dimension of 300, which is the same as the text features produced by word2vec.

\subsubsection{Testing Modality}\label{sec:testmod}
We report the experimental results for three settings based on the testing modality, i.e., test data point has a) only audio or b) only video or c) both data.  When the query data point contains only audio or only video data, we can directly calculate the distance of the query data point from all the class label embeddings using the audio or video embedding, respectively. Using this distance, we can find the nearest class label embedding, which is the predicted class.  

When the query data point contains both audio and video data, we calculate the distance from all the class label embeddings for both audio and video data. We compute the average of the audio and video distances and use the mean distance to find the nearest class label text embedding (AVGZSLNet (eq wt)). We show experimentally that AVGZSLNet (eq wt) significantly outperforms all the base-lines and the compared methods.  

The authors in \cite{parida2020coordinated} perform a weighted addition of the audio and video distances and use it to find the nearest neighbor class in the text embedding space. They use an attention module to learn the attention weights for the audio and video distances. We provide an additional result using this attention mechanism (AVGZSLNet (w/ attn)). We would like to point out that our method, AVGZSLNet (eq wt), outperforms the state-of-the-art attention-based method CJME, even though we do not use attention.

\subsubsection{Base-lines and Compared methods}
We compare our generalized zero-shot classification and retrieval results with several base-line models, namely, the audio-only model, the video-only model, the audio-video model,  and the concatenation model. The audio-only model consists of a single projection/embedding network projecting audio features to the embedding space, and it trains to push the audio embeddings closer to the corresponding class label embedding. The video-only model is similar, but it uses only video data. The audio-video concatenation model also trains only one projection network, but it uses the audio-video concatenated features as input. It needs both audio and video data to perform training or testing. Therefore, this model cannot work when only audio or video modality is present. For the retrieval experiments, we also report the results for a base-line referred to as the pre-trained model. This model uses the features produced by the pre-trained networks (used to extract the initial features for our method) to perform retrieval. It can use only audio or video features at a time.

We also compare our results with other methods in this setting. Coordinated Joint Multimodal Embeddings \cite{parida2020coordinated} makes use of pair-wise cross-modal similarity loss between audio and video embeddings of the same data-point and triplet loss with text embedding as anchor and audio/video embeddings as the positive and negative inputs. Generalized Canonical Correlation Analysis \cite{kettenring1971canonical} is a standard method for maximizing the correlation between example pair-wise data. In this setting, GCCA is used to maximize the correlation between audio, video, and text for every data-point. We report the retrieval results for GCCA. We also compare our method with zero-shot learning approaches, namely, CONSE, DEVISE, SAE, ESZSL, ALE. The results for the audio-only, video-only, pre-trained, CONSE, DEVISE, SAE, ESZSL, ALE, CJME models are the same as reported in \cite{parida2020coordinated}. 

\begin{table}
\centering
\resizebox{\columnwidth}{!}{
\begin{tabular}{c|c|ccc}
\hline
Model & Test Modality & S & U & HM \\
\hline \hline 
audio only model \cite{parida2020coordinated}         & audio & 28.35 & 18.35 & 22.22 \\
CJME \cite{parida2020coordinated}          & audio & 25.58  & 20.30 & 22.64 \\
AVGZSLNet (ours)      & audio & 33.29 & 20.43 & \textbf{25.32} \\
\hline
video only model \cite{parida2020coordinated}         & video & 43.27 & 27.11 & 33.34 \\
CONSE~\cite{parida2020coordinated,norouzi2013zero}   & video &48.50 & 19.60 & 27.90  \\
DEVISE~\cite{parida2020coordinated,frome2013devise}  & video &39.80 & 26.00 & 31.50  \\
SAE~\cite{parida2020coordinated,kodirov2017semantic}   & video &29.30 & 19.30 & 23.20 \\
ESZSL~\cite{parida2020coordinated,romera2015embarrassingly} & video & 33.80 & 19.00 & 24.30 \\
ALE~\cite{parida2020coordinated,akata2015label}  & video & 47.90 & 25.20 & 33.00 \\
CJME \cite{parida2020coordinated} & video & 41.53 & 28.76 & 33.99 \\
AVGZSLNet (ours)     & video & 48.05 & 30.71 & \textbf{37.47} \\
\hline
audio-video concat model & both & 45.83 & 27.91 & 34.70 \\
CJME (eq wt) \cite{parida2020coordinated}   & both   & 30.29 & 31.30 & 30.79 \\
AVGZSLNet (ours) (eq wt)     & both & 46.04 & 33.80 & \textbf{38.98} \\
\hline
CJME \cite{parida2020coordinated} (w/ attn) & both & 41.07 & 29.58 & 34.39\\
AVGZSLNet (ours) (w/ attn)    & both & 50.96 & 32.73 & \textbf{39.86} \\
\hline
\end{tabular}
}
\caption{Generalized zero-shot classification mean class accuracy (\% mAcc) achieved with audio or video or both (audio and video) data during testing.}
\label{tab:zsc}
\vspace{-15pt}
\end{table}
\subsection{Metrics}
We use the mean class accuracy (\% mAcc) metric for classification and the mean average precision (\% mAP) metric for retrieval \cite{xian2018zero}. We perform classification/retrieval for all classes and then report the performance for seen (S) and unseen (U) classes.
\begin{equation}
    HM = \frac{2\times U \times S}{U + S}
\end{equation}

We focus on the harmonic mean (HM) of the performances on seen and unseen classes and try to achieve higher HM for our method as it is not biased towards the seen or unseen classes. This is a general practice in generalized zero-shot learning classification. We have used the same metric for both classification and retrieval.

\subsection{Generalized Zero-Shot Classification}

Table \ref{tab:zsc} reports the performance of our method for generalized zero-shot classification. When only audio or video data is present at test time, our model significantly outperforms the audio-only and video-only base-line models. This shows that training on both audio and video has helped our model to generalize better than training on only audio or video. Our model also significantly outperforms CJME by an absolute margin of 2.68\% and 3.48\% in the HM metric for the cases when the test modality is only audio and only video, respectively.

When both audio and video modalities are present in the query data points, AVGZSLNet (eq wt) outperforms CJME (eq wt) by an absolute margin of 8.19\% (HM) and CJME (w/ attn) by an absolute margin of 4.49\% (HM). AVGZSLNet (eq wt) also outperforms the audio-video concatenation model by an absolute margin of 4.28\% (HM). AVGZSLNet (w/ attn) outperforms CJME (w/ attn) by a margin of 5.47\% in the HM value. The results clearly indicate that our method AVGZSLNet (eq wt) significantly outperforms CJME (w/ attn) without even using attention. This shows the effectiveness of our method.

\begin{figure*}[t]
  \centering
  \includegraphics[width=0.8\linewidth]{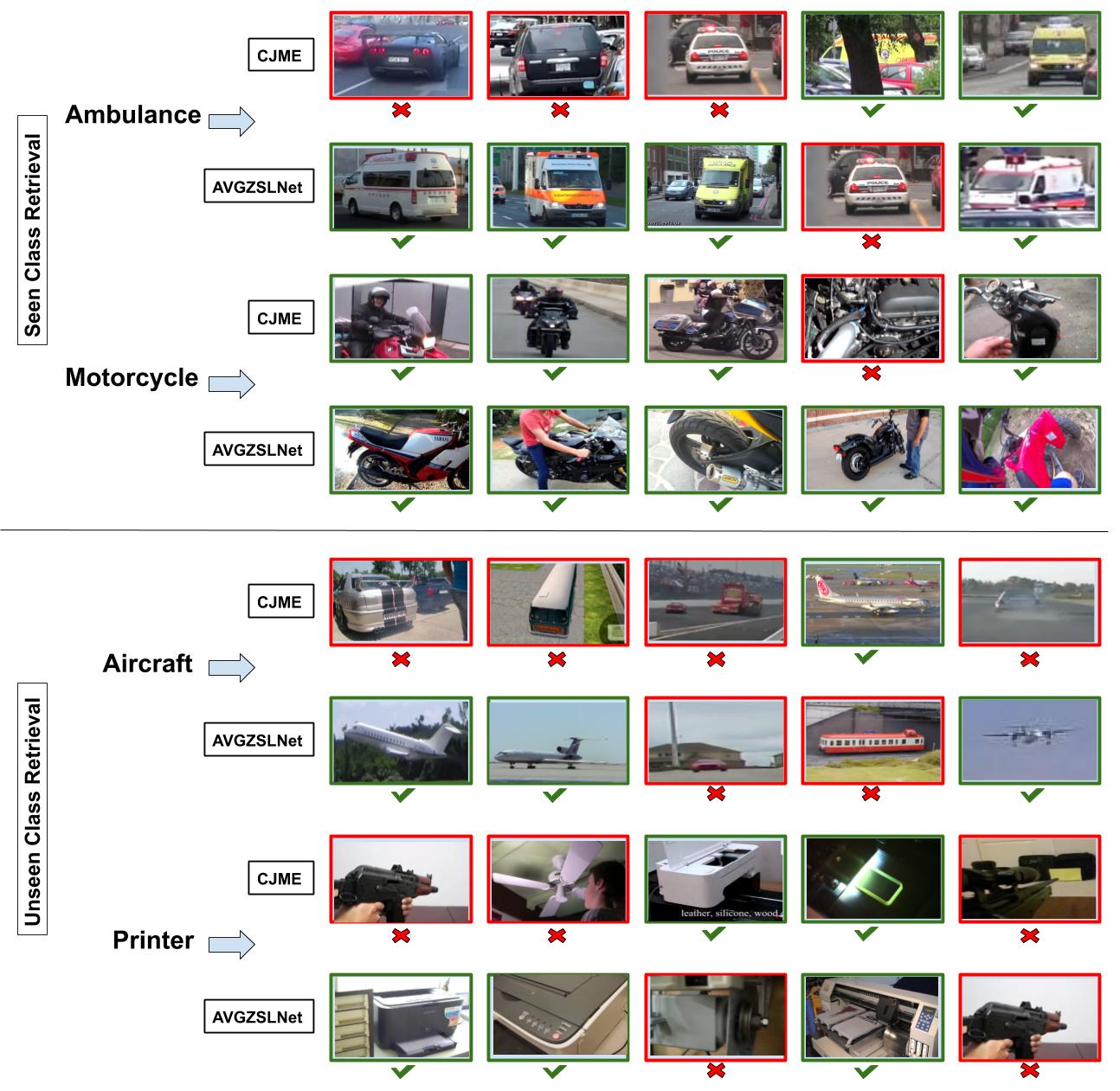}
  \caption{Generalized Zero-Shot Retrieval Qualitative Results.}
  \label{fig:qual}
  \vspace{-15pt}
\end{figure*}

\subsection{Generalized Zero-Shot Retrieval}
\begin{table}
\centering
\resizebox{\columnwidth}{!}{
\begin{tabular}{c|c|ccc}
\hline
Model & Test & S & U & HM \\
\hline \hline
pre-trained model \cite{parida2020coordinated} & T $\rightarrow$ A & 3.83 &  1.66 & 2.32 \\
GCCA \cite{parida2020coordinated,kettenring1971canonical} & T $\rightarrow$ A & 49.84 &  2.39 & 4.56 \\
audio only model \cite{parida2020coordinated}  & T $\rightarrow$ A & 43.16 & 3.34 & 6.20 \\
CJME \cite{parida2020coordinated}  & T $\rightarrow$ A & 48.24  &  3.32  &  6.21 \\
AVGZSLNet (ours)  & T $\rightarrow$ A & 51.55  &  3.85  &  \textbf{7.17 }\\
\hline
pre-trained model \cite{parida2020coordinated} & T $\rightarrow$ V & 3.83 & 2.53 & 3.05 \\
GCCA \cite{parida2020coordinated,kettenring1971canonical} & T $\rightarrow$ V & 57.67 &  3.54 & 6.67 \\
video only model \cite{parida2020coordinated}  & T $\rightarrow$ V & 48.62 & 5.25 & 9.47 \\
CJME \cite{parida2020coordinated} & T $\rightarrow$ V & 59.39  & 5.55  & 10.15 \\
AVGZSLNet (ours)  & T $\rightarrow$ V & 62.20  & 6.39 &  \textbf{11.60} \\
\hline
CJME (eq wt) \cite{parida2020coordinated} & T $\rightarrow$ AV & 65.45 & 5.40 & 9.97\\
AVGZSLNet (ours) (eq wt)  & T $\rightarrow$ AV & 72.25  &  6.91  &  \textbf{12.61} \\
\hline
CJME (w/ attn) \cite{parida2020coordinated} & T $\rightarrow$ AV & 62.97 & 5.67 & 10.41\\
AVGZSLNet (ours) (w/ attn)  & T $\rightarrow$ AV & 68.94 & 7.08 &  \textbf{12.84} \\
\hline
\end{tabular}
}
\caption{Generalized zero-shot retrieval mean average precision (\% mAP) achieved with only audio, only video, and both (audio and video) data during testing.}
\label{tab:zsr}
\vspace{-15pt}
\end{table}

 Table \ref{tab:zsr} shows generalized zero-shot retrieval results from the class label text, i.e., retrieving data points from the dataset in the order of how close they are to the class label text in the embedding space. The results for unseen classes for all cases are very low. We can attribute this observation to the bias of the trained models towards seen classes in generalized zero-shot learning. We can address this problem in classification by reducing the scores of the seen classes. However, we cannot make this correction here as there is no concept of class or scores in retrieval. We use HM as a metric for zero-shot retrieval for a fair comparison with \cite{parida2020coordinated}. Other zero-shot retrieval approaches have used a mean of seen and unseen class accuracy as the metric. Such a metric will give higher results but will not give a clear picture of seen and unseen accuracy.

From Table \ref{tab:zsr}, we can see that when performing retrieval from class label text embedding using only audio or only video embedding, our method AVGZSLNet performs better than CJME, the base-line audio-only model, the pre-trained model, and GCCA. 

When both audio and video data is present for each data-point, AVGZSLNet (eq wt) outperforms CJME (eq wt) by a margin of 2.64\% in HM value. It should be noted again that our method AVGZSLNet (eq wt) significantly outperforms CJME (w/ attn) without even using attention.  AVGZSLNet (w/ attn) outperforms CJME (w/ attn) by a margin of 2.43\% in HM value. We provide some qualitative results in Fig \ref{fig:qual} comparing our method and CJME for the generalized zero-shot retrieval task on the AudioSetZSL dataset. The retrieval experiments use the class label texts as anchor.
 
 \begin{table}[t]
 \centering
\resizebox{\columnwidth}{!}{
 \begin{tabular}{c||c|c|c||ccc}
 \hline
 $L_{CT}$ &$L_{REC}$ & $L_{CTA}$ & $L_{CTV}$ & S & U & HM \\
 \hline 
  $Y$ & $N$ &$Y$ & $Y$  & 42.76 & 32.88 & 37.17  \\
 $Y$&$Y$ &$N$ & $Y$  &  46.40  &  31.84  &  37.77\\
 $Y$&$Y$ &$Y$ & $N$  & 45.87  & 31.54  &  37.38   \\
  $Y$ &$Y$ & $Y$ & $Y$  &  46.04 & 33.80 & \textbf{38.98}   \\

\hline
 \end{tabular}
 }
 \caption{Ablation study to verify the contribution of $L_{REC}$, $L_{CTA}$ and $L_{CTV}$ components of $L_{CMD}$ on AVGZSLNet (eq wt) for generalized zero-shot classification (\% mAcc).}
 \label{tab:ablcmd}
\vspace{-15pt}
 \end{table}

\section{Ablations}

\subsection{Significance of the components of $L_{CMD}$}
We perform ablation experiments to verify the contribution of $L_{REC}$, $L_{CTA}$ and $L_{CTV}$ components of our cross-modal decoder loss ($L_{CMD}$). From Table \ref{tab:ablcmd}, we observe that removing any of these three components impacts the network performance negatively. Therefore, $L_{REC}$, $L_{CTA}$, and $L_{CTV}$ are important components, and consequently, cross-modal decoder loss is also significant.
  \begin{table}[t]
 \centering
\resizebox{\columnwidth}{!}{
 \begin{tabular}{c||c|c|c|c||ccc}
 \hline
 $L_{CMD}$ &$L_{TA}$ & $L_{AT}$ & $L_{TV}$ & $L_{VT}$ & S & U & HM \\
 \hline 
  $Y$ & $N$ &$Y$ & $Y$ & $Y$  &  48.91 & 27.65 & 35.33  \\
 $Y$&$Y$ &$N$ & $Y$ & $Y$ &  44.28  &  29.96  &  35.74 \\
 $Y$ &$Y$ &$Y$ & $N$ & $Y$ & 51.27  &  28.54  &  36.67  \\
 $Y$&$Y$ &$Y$ & $Y$ & $N$ & 45.91  &  27.90  &  34.71  \\
  $Y$ &$Y$ & $Y$  &$Y$ & $Y$ &  46.04 & 33.80 & \textbf{38.98}   \\
\hline
 \end{tabular}
 }
 \caption{Ablation study to verify the contribution of $L_{TA}$, $L_{AT}$, $L_{TV}$ and $L_{VT}$ components of $L_{CT}$ on AVGZSLNet (eq wt) for generalized zero-shot classification (\% mAcc).}
 \label{tab:abltrip}
\vspace{-15pt}
 \end{table}

\begin{figure*}[htb]
     \centering
     \scalebox{0.85}{
     \begin{subfigure}[b]{0.49\textwidth}
         \centering
         \includegraphics[width=\textwidth]{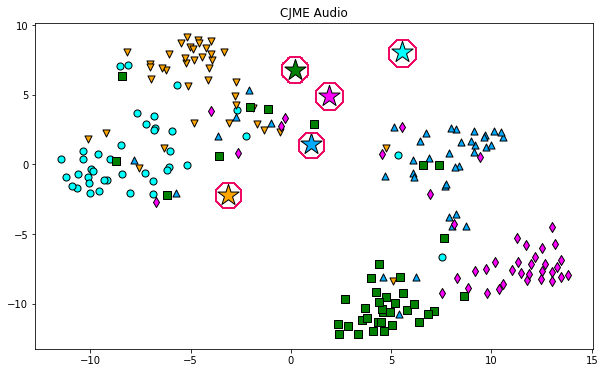}
         \label{fig:audiocjme}
     \end{subfigure}
     \hfill
     \begin{subfigure}[b]{0.49\textwidth}
         \centering
         \includegraphics[width=\textwidth]{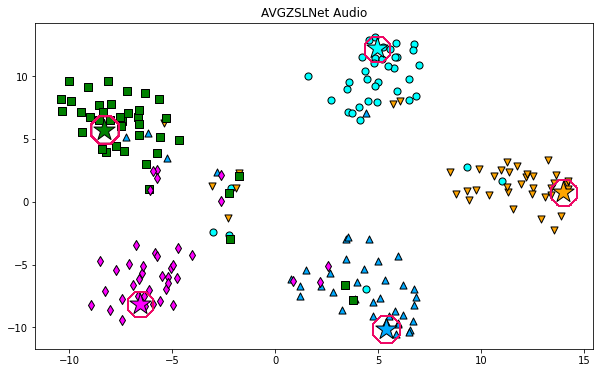}
         \label{fig:audioavsl}
     \end{subfigure}
     }
    \scalebox{0.85}{
     \begin{subfigure}[b]{0.49\textwidth}
         \centering
         \includegraphics[width=\textwidth]{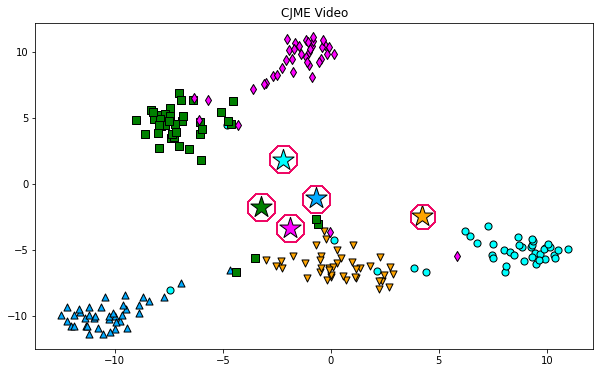}
         \label{fig:videocjme}
     \end{subfigure}
     \hfill
     \begin{subfigure}[b]{0.49\textwidth}
         \centering
         \includegraphics[width=\textwidth]{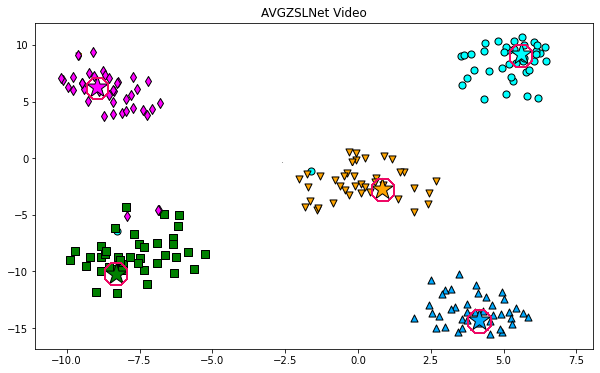}
         \label{fig:videoavsl}
     \end{subfigure}
     }
        \caption{t-SNE plots of video/audio embeddings along with the class label text embeddings of the test samples of 5 randomly chosen seen classes of AudioSetZSL and each class is represented by one color. Stars indicate the class label text embedding of each class. Both audio and video embeddings, produced by AVGZSLNet are significantly closer to the respective class label text embedding (same color) as compared to CJME. For AVGZSLNet, the stars (class label text embeddings) are surrounded by audio/video embeddings of the same class.}
        \label{fig:tsne}
        \vspace{-10pt}
\end{figure*}

\subsection{Significance of the components of $L_{CT}$}
We perform ablation experiments to verify the contribution of $L_{AT}$, $L_{TA}$, $L_{VT}$ and $L_{TV}$ components of our composite triplet loss ($L_{CT}$). The results in Table \ref{tab:abltrip} indicate that if we drop any of these four components, then the network performance drops significantly. Therefore, $L_{AT}$, $L_{TA}$, $L_{VT}$, and $L_{TV}$ are important components, and consequently, composite triplet loss is also a significant component of our method.

\section{Visualizations}

We provide t-SNE plots in Fig.~\ref{fig:tsne} of video and audio embeddings along with the class label text embeddings of 40 randomly chosen query samples from 5 randomly chosen seen classes of AudioSetZSL for CJME and AVGZSLNet. Please note that we used the same query data points for both AVGZSLNet and CJME models for a fair comparison. The t-SNE plots for both audio and video embeddings show that AVGZSLNet audio/video embeddings are significantly closer to the class label text embedding as compared to CJME. Therefore, CMD loss and CT loss have helped our projection networks to produce embeddings that are closer to the text embedding of the same class.

\section{Statistical Significance}
\begin{figure}[t]
  \centering
  \includegraphics[width=0.25\textwidth]{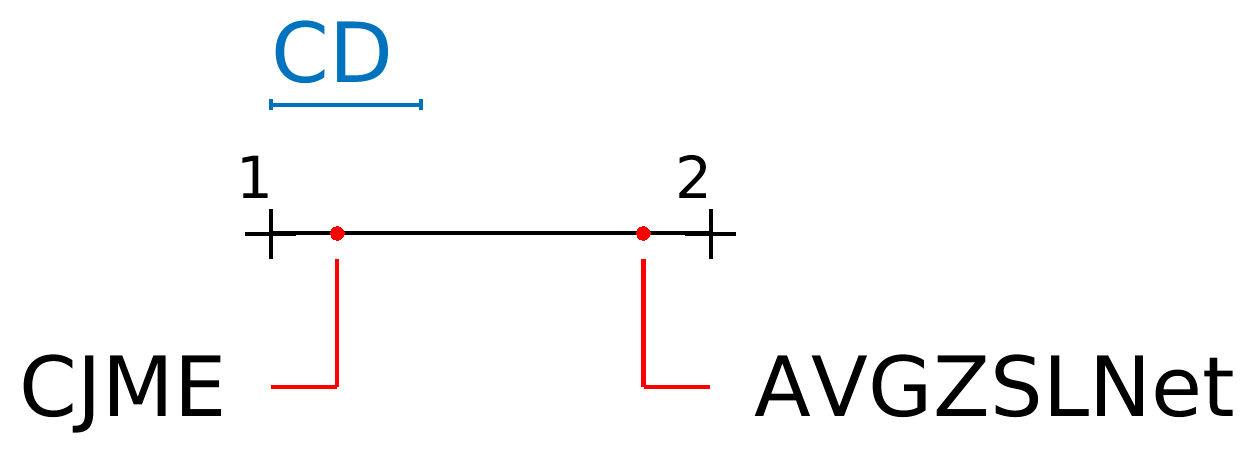}
  \caption{Statistical significance test to show that AVGZSLNet is statistically different from CJME with a significance level of 0.05.}
  \label{fig:statsig}
  \vspace{-15pt}
\end{figure}

We studied the statistical significance \cite{demvsar2006statistical} for our method AVGZSL against CJME. If the difference in the rank of the two methods lies within Critical Difference, then they are not significantly different. Fig.~\ref{fig:statsig} visualizes the post hoc analysis using the CD diagram for generalized zero-shot classification. We observe in Fig.~\ref{fig:statsig} that the statistical difference between AVGZSLNet and CJME is almost twice the Critical Difference (CD=0.3412). Therefore, AVGZSLNet is statistically different from CJME. 

\section{Conclusion}
We propose a method for audio-visual generalized zero-shot learning in a multi-modal setting using a combination of cross-modal decoder loss and composite triplet loss to improve the performance. Training on these losses helps the audio and video embeddings to move closer to the text embeddings of the corresponding class labels. We empirically show that our method outperforms the state-of-the-art method on audio-visual generalized zero-shot classification and retrieval. Through ablation experiments, we validate the choice of the losses that we propose.

{\small
\bibliographystyle{ieee_fullname}
\bibliography{egbib}
}

\end{document}